\documentclass[a4paper, 10pt, conference]{ieeeconf}      % Use this line for a4
\IEEEoverridecommandlockouts                              % This command is only
\usepackage{graphics,lipsum} % for pdf, bitmapped graphics files
\usepackage{epsfig} % for postscript graphics files
\usepackage{cuted,subcaption}
\usepackage{amsmath} % assumes amsmath package installed
\usepackage{amssymb}  % assumes amsmath package installed

\title{\LARGE \bf {A reinforcement learning algorithm for building collaboration in multi-agent systems}}

\author{ \parbox{3 in}{\centering Mehmet E. Aydin*
        \thanks{*Corresponding author}\\
        Department of Computer Science and Creative Technologies\\
        University of the West of England\\
        Frenchay Campus, Bristol, UK\\
        {\tt\small mehmet.aydin@uwe.ac.uk}}
        \hspace*{ 0.5 in}
        \parbox{3 in}{ \centering Ryan Fellows\\
       Department of Computer Science and Creative Technologies\\
        University of the West of England\\
        Frenchay Campus, Bristol, UK\\
        {\tt\small ryan.fellows@uwe.ac.uk}}
}

\begin{document}

\maketitle
\thispagestyle{empty}
\pagestyle{empty}

\begin{abstract}

This paper presents a proof-of concept study for demonstrating the viability of building collaboration among multiple agents through standard Q learning algorithm embedded in particle swarm optimisation. Collaboration is formulated to be achieved among the agents via some sort competition, where the agents are expected to balance their action in such a way that none of them drifts away of the team and none intervene any fellow neighbours territory. Particles are devised with Q learning algorithm for self training to learn how to act as members of a swarm and how to produce collaborative/collective behaviours.  The produced results are supportive to the algorithmic structures suggesting that a substantive collaboration can be build via proposed learning algorithm.

\end{abstract}

\section{INTRODUCTION}

Cutting-edge technologies facilitates the daily-life of individuals and societies with more opportunities to overcome challenging issues continuously introducing new smart gadgets day-in day-out. These astonishing technologies introduce changes with use smart sensors in most of the time, which places a crucial role in our daily life as they are literally everywhere any more. Internet of Things (IoT) is one of key technologies to organise smart sensors in order to facilitate living environments with more and more services such as Smart homes and cities, highly-efficient engineering products, crews/swarms of robots etc. A particular example can be a swarm of unmanned aerial vehicles (UAVs) which are teamed up to collect information from disaster areas to predict/discover and help identify the impact of damage and the level of human suffering.  This is due to the fact that information collection plays a very crucial role in disaster management, where the decisions are required to be done timely and based on correct and up-to-date information. Swarms of UAVs can be devised for this purposes, which are expected to remain inter-connected all the time to deliver the duties collaboratively~\cite{Aydin_et_al_11}. Obviously, this is a typical implementation area of IoT, where smart sensors and tiny devices, which are drones (UAVs) in this case, require efficient and robust settings and configuration. However, an efficiently exploring swarm is not easy to design and run due to various practical issues such as energy limitations. This paper introduces a novel learning algorithm to train individual devices to make smartly behaving and collaborating entities. 

Multi-agent systems (MAS) is an up-to-date artificial intelligence paradigm , which attracts much attention for modeling intelligent solutions in rather a distributed form. It imposes formulating limited capacity items as proactive and smart entities, which autonomously act and accumulate experience to exploit ahead in fulfilling duties more and more efficiently. In this way, a more comprehensive and collective intelligence can be achieved. This paradigm has proved success so many times in a wider problem solving horizon~\cite{Aydin_10, Kazemi_09,Aydin_07,Kouider_12}. This proves that developing IoT models using MAS paradigm will produce a substantial benefit and efficiency. However, building a collaboration among multiple agents remains challenging since MAS studies have not reached to a sufficient level of maturity due to the difficulty in the nature of the problem.  The remaining parts of this paper introduce a novel learning algorithm implemented for multiple agent models, where a collaboration is aimed to be constructed among the participating agents via the introduced algorithm. It is a reinforcement learning algorithm, which best fits real-time learning cases, and dynamically changing environments. The individual agents are expected to learn from past experiences for which how to stay interconnected and remain as a crew to collectively fulfill the duties without wasting resources. The latter purpose enforces the individual agents to compete in achieving higher rewards through out of the entire process, which makes the study further important since collaboration has to be achieved while competing. Previously, competition-based collective learning algorithm has been attempted with learning classifier systems for modelling social behaviours~\cite{Hercog_13}. Although there are many other studies conducted for collective learning of multi-agents with Q learning~\cite{Foerster_16, Panait_05}, the proposed algorithm implements a competition-based collective learning algorithm extending Q learning with the notion of individuals and their positions in particle swarm optimisation (PSO) algorithm, which ends up as Q learning embedded in PSO. 

The rest of the paper consists of the following structure; the background and literature review is presented in Section~\ref{backg}, the proposed reinforcement learning algorithm is introduced in Section~\ref{learning_birds}, the implementation of the algorithm for scanning fields is elaborated in Section~\ref{implementation}, experimental results and discussions are detailed in Section~\ref{results} and finally conclusions in Section~\ref{conclusions}. 

\section{Background}\label{backg}
\subsection{Swarm Intelligence}\label{si}
Swarm intelligence is referred to artificial intelligence (AI) systems where an intelligent behaviour can emerge as the the  self-organised outcome of a collection of simple entities such as agents, organisms or individuals. Simple organisms that live in colonies; such as ants, bees, bird flocks etc. have long fascinated many people for their collective intelligence and emergent behaviours that is manifested in many of activities they do. A population of such simple entities can interact with each other as well as with their environment without using any set of instruction(s) to proceed, and compose a swarm intelligence system~\cite{Kennedy_01}, .

The swarm intelligence approaches are to reveal the collective behaviour of social insects in performing specific duties; it is about modelling the behaviour of those social insects and use these models as a basis upon which varieties of artificial entities can be developed. In such a way, the problems can be solved by models that exploit the problem solving capabilities of social insects. The motivation is to model the simple behaviours of individuals and the local interactions with the environment and neighbouring individuals, in order to obtain more complex behaviours that can be used to solve complex problems, mostly optimisation problems \cite{Colorni_et_al_94},~\cite{Tasgetiren_07}).

\subsection{Reinforcement Learning}\label{rl}
Reinforcement learning (RL) is a class of learning in which unsupervised learning rules work alongside with a reinforcement mechanism to reward an agent based on its action selection activity to respond the stimulus from its own environment. It can be also called as semi-supervised learning since it receives a reinforcement point, either immediate or delayed, fed back from the environment. Let $\Lambda$ be an agent works in environment $E$, which stimulates $\Lambda$ with its state $s\in S$, where $S$ is the finite set of states of $E$. The agent $\Lambda$ will evaluate this perceived state and make a decision to select an action $a\in A$, where $A$ is the finite set of actions that an agent can take. Meanwhile, the reinforcement mechanism, may also be called as reward function, assesses the action, $a$, taken by $\Lambda$ in response to state $s$ and  produces reward $r$ to feed back to $\Lambda$. Here, the ultimate aim of the agent $\Lambda$ is to maximize its accumulated reward by the end of the learning period/process, as in the following expression:
\begin{equation}\label{reward}
    \max \mathbf{R}  =  \sum_{i=1}^\infty r_i
\end{equation}
where $\infty$ is practically replaced with a finite number such as $I$ to be the total number of learning iterations. Although an agent is theoretically expected to function forever, it usually works for a predefined time period as a matter of practicality.

There are various reinforcement learning methods developed with various properties. Among these, Q Learning~\cite{Watkins_92}, \cite{Tsitsiklis_94}, TD Learning~\cite{Bradtke_96},\cite{Tesauro_92}, learning classifier systems~\cite{Bull_05}, \cite{Bull_et_al_05} etc are well know reinforcement learning approaches.

\subsection{Collaboration in multi agent systems}
\label{mas_collab} Multi agent systems (MAS) are well- known and relatively mature distributed collective intelligence approaches with which a set of proactive agents act individually for solving the problems in collaboration ~\cite{Ayhan_13}. The main theme is to team up intelligent autonomous entities for solving the problems in harmony and composing a certain level of coordination to help individual agents act proactively and efficiently to contribute and collaborate in problem solving process demonstrating individual intelligence capacity~\cite{Aydin_10}. It is useful to note that the main properties of MAS (i.e. autonomy, responsiveness, redundancy, and distributed approach) facilitate success in MAS applications, which result in a good record in implementations within many research fields including production planning, scheduling and control~\cite{Mohebbi_12}, engineering design, and process planning~\cite{Ayhan_13}.

The concept of metaheuristic agents has recently been identified to describe a particular implementation of multi agent systems devised to tackle hard optimisation problems. The idea is to build up teams of individual agents equipped with metaheuristic problem solvers aiming to solve hard and large-scale problems with distributed and collaborative intelligent search skills.  In the literature, few multi agent systems implementing metaheuristics are introduced and overviewed with respect to their performances ~\cite{Aydin_07,Hammami_05} while it is known that metaheuristic approaches are, by large, used as standalone applications.

Researchers are conscious on that solving complex and large problems with distributed approaches remains as a challenging issue due to the fact that there is not a productive method to commonly use for organising distributed intelligence (agents in this case) for a high efficiency.  ~\cite{Kolp_06, Panait_05, Vazquez_05}. In fact, the performances of multi- agent systems including metaheuristic teams significantly depends on the quality of collaboration~\cite{Aydin_07}. Swarm intelligence-based agent collaboration is suggested in~\cite{Aydin_10}, while the persistence of this challenging issue is reflected in a number of recent studies including~\cite{Gath_15} and \cite{Dong_16}, where ~\cite{Gath_15} introduces auction-based consensus among the agents while \cite{Dong_16} studies theoretical bases of agent collaboration through mathematical foundations.  

\section{Modelling with Swarms of Learning Agents}\label{learning_birds}
\subsection{Q learning}\label{q_learn}
Q learning is a reinforcement learning algorithm that is developed based on temporal-difference handled with asynchronous dynamic programming. It provides rewards for agents with the capability of learning to act optimally in Markovian domains by experiencing the consequences of actions, without requiring them to build map of the respective domain \cite{Watkins_89}. The main idea behind Q learning is to use a single data structure called the utility
function ($Q(x,a)$). That is the utility of performing action $a$ in state $x$ \cite{Watkins_92}. Throughout the whole learning process, this algorithm updates the value of $Q(x,a)$ using $x$, $a$, $r$, $y$ tuples per step, where $r$ represents the reinforcement signal (payoff) of the  environment and $y$ represents the new state which is obtained as the consequence of executing action $a$ in state $x$. Both $x$ and $y$ are elements
of the set of states ($S$) and $a$ is an element of the set of actions ($A$).  $Q(x,a)$ is defined as:-
\begin{equation}
\mathbf{Q}:S\times A\;\longrightarrow \Re
\end{equation}
and determined as:-
\begin{equation}
Q(x,a)=E(r+\gamma e(y)|x,a)
\end{equation}
where $\gamma$ is a discounted constant value within the interval of [0,1] as described according to the domain and $e(y)$ is the expected value of $y$ defined as:
\begin{equation}
e(y) = \max \{Q(y,a)\}\qquad for \quad\forall a\in A 
%\hspace{0.35cm}for\hspace{0.1cm} \forall a\in A 
\end{equation}

The learning procedure first initialises the $Q$ values to $0$ for each  action. It then repeats the following procedure. The action with the maximum $Q$ value is selected and activated. Corresponding $Q$ value of that action is then updated using the following equation (updating rule):-
\begin{equation}
 Q^{t+1}(x,a)=Q^{t}(x,a)+\beta (r+\gamma e(y)-Q^{t}(x,a))
\end{equation}
where $Q^t (x,a)$ and  $Q^{t+1} (x,a)$ are the old and the new $Q$ values of action $a$ in state $x$, respectively. $\beta$ is the learning coefficient changing in  [0,1]  interval. This iterative process ends when an acceptable level of learning is achieved or a stopping criterion is satisfied. For more information see Sutton and Barto \cite{Sutton_98}.

\subsection{Particle swarm optimisation (PSO)} \label{sec:pso}
PSO is a population-based optimization technique inspired of social behaviour of bird flocking and fish schooling. PSO inventors have implemented such natural processes to solve the optimization problems in which each single solution, called a particle, joins the other individuals to make up a swarm (population) for exploring within the search space. Each particle has a fitness value calculated by a fitness function, and a velocity of moving towards the optimum. All particles search across the problem space following the particle nearest to the optimum. PSO starts with initial population of solutions, which is updated iteration-by-iteration. A basic PSO algorithm builds each particle based on, mainly, two key vectors; position vector,
$\mathbf{x_i(t)} = \{x_{i,1}(t),...,x_{i,n}(t)\}$, and velocity vector $\mathbf{v_i(t)} = \{v_{i,1}(t),...,v_{i,n}(t)\}$, where $x_{i,k}(t)$, is the position value of the $i^{th}$ particle with respect to the $k^{th}$ dimension $(k=1,2,3,..,n)$ at iteration $t$, and $v_{i,k}(t)$ is the velocity value of the $i^{th}$ particle with respect to the $k^{th}$ dimension at iteration $t$. The initial values, $\mathbf{x_i(0)}$ and $\mathbf{v_i(0)}$, are given by

\begin{eqnarray}
\label{position update}
x_{i,k}(0)& = &x_{min}+(x_{max}-x_{min})\times r_1,\\
v_{i,k}(0)& = &v_{min}+(v_{max}-v_{min})\times r_2,%\label{velocity update}
\end{eqnarray}
where $x_{min}, x_{max}, v_{min}, v_{max}$ are lower and upper limits of the ranges of position and velocity values, respectively, and finally, $r_1$ and $r_2$ are uniform random numbers within $[0,1]$.  Since both vectors are continuous, the original PSO algorithm can straightforwardly be used for continuous optimization problems. However, if the problem is combinatorial, a discrete version of PSO needs to be implemented. Once a solution is obtained, the quality of that solution is measured with a cost function denoted with $f_i$, where $f_i:\mathbf{x_i(t)}\longrightarrow \Re$.

For each particle in the swarm, a personal best, $\mathbf{y_i(t)}=\{y_{i,1}(t),...,y_{i,n}(t)\}$, is defined, where $y_{i,k}(t)$ denotes the position of the $i^{th}$ personal best with respect to the $k^{th}$ dimension at iteration $t$. The personal bests are equal to the corresponding initial position vector at the beginning. Then, in every generation, they are updated based on the solution quality. Regarding the objective function, $f_i$, the fitness values for the personal best of the $i^{th}$ particle, $\mathbf{y_i(t)}$, is denoted by $f^{y}_{i}(t)$ and updated whenever $f^{y}_{i}(t+1) \prec f^{y}_{i}(t)$, where $t$ stands for iteration and $\prec$ corresponds to the logical operator, which becomes $<$ or $>$ for minimization or maximization problems respectively.

On the other hand, a global best, which is the best particle within the whole swarm is defined and selected among the personal bests, $\mathbf{y(t)}$, and denoted with $ \mathbf{g(t)}=\{g_1(t),...,g_n(t)\}$. The fitness of the global best, $f_g(t)$, can be obtained using:
\begin{eqnarray}
f_g(t) & = & \mathbf{opt}_{i\in{N}}\{f^{y}_i(t)\} 
\label{fg}
\end{eqnarray}
where $\mathbf{opt}$ becomes $\min $ or $\max$ depending on the type  of optimization. Afterwards, the velocity of each particle is updated based on its personal best, $\mathbf{y_i(t)}$ and the global best, $\mathbf{g(t)}$ using the following updating rule:

\begin{eqnarray}\label{update}
  \mathbf{v_i}(t+1) &=& \delta w_t \Delta\mathbf{v_i}(t) \\
 \Delta\mathbf{v_i} &=& \scalebox{0.85}[1] {$c_1r_1(\mathbf{y_i}(t)-\mathbf{x_i}(t))+c_2r_2(\mathbf{g}(t)-\mathbf{x_i}(t))$}
\end{eqnarray}

where $w$  is the inertia weight used to control the impact of the previous velocities on the current one, which is decremented by $\beta$, decrement factor, via $w_{t+1}=w_{t}\times \beta $, $\delta$ is constriction factor which keeps the effects of the randomized weight within the certain range. In addition, $r_1$ and $r_2$ are random numbers in [0,1] and $c_1$ and $c_2$ are the learning factors, which are also called social and cognitive
parameters. The next step is to update the positions with:
\begin{eqnarray}\label{final_update}
  \mathbf{x_i}(t+1) &=& \mathbf{x_i}(t)+\mathbf{v_i}(t+1)
\end{eqnarray}
for continues problem domains. On the other hand, since discrete problems cannot be solved in the same way of continuous problems, various discrete PSO algorithms have been proposed. Among these, Kennedy and Eberhart \cite{Eberhart_95} have proposed the most used one, which mainly creates binary position vector based on velocities as follows:
\begin{equation}
\mathbf{x_i}(t+1)  =  \frac{1}{e^{\mathbf{v_i}(t+1)}}.
\label{final_update_1}
\end{equation}

After getting position values updated for all particles, the corresponding solutions with their fitness values are calculated so as to start a new iteration if the predetermined stopping criterion is not satisfied. For further information, \cite{Kennedy_97} and \cite{Tasgetiren_07} can be seen.

\subsection{Swarms of Learning Agents} \label{swarm_learning_agents}
PSO is one of very well know swarm intelligence algorithms used to develop collective behaviours and intelligence inspiring of bird flocks. Although it has a good record of success, learning capability remains an important aspect to be developed further for an improved intelligence. There are few studies investigating the hybridisation of reinforcement learning algorithms, especially Q Learning algorithm implemented for particular applications~\cite{Claus_98}, \cite{Kok_04}, \cite{Meng_08}. Likewise, Q Learning algorithm has been implemented by various studies to develop coordination of multi agent systems ~\cite{Iima_09}. However, PSO has not been integrated with Q Learning in order to make each particle within the swarm towards learning for collaboration.

For the purpose of training the particles of the swarm to behave in harmony within its neighbourhood, we propose use of Q Learning algorithm in building intelligent search behaviour of each individual. A Q Learning algorithm is embedded in PSO in a way that the position vectors, $\mathbf{x_i}$, is updated subject to a well-designed implementation of Q learning to adaptively control the behaviour of the individuals towards collective behaviours, where all individual members of the swarm collectively and intelligently contribute. Hence, we revised PSO, first, with ignoring the use of velocity vector, $\mathbf{v_i}$, so as to save time and energy relaying on the fact that the position vector, $\mathbf{x_i}$, inherently contains $\mathbf{v_i}$, and does not necessitate its use~\cite{Poli_07}, \cite{Aydin_et_al_09}. Secondly, the update rule of the position vectors, $\mathbf{x_i}$, (Eq:~\ref{final_update}) is revised as follows:
\begin{eqnarray}
 \mathbf{x_i}(t+1) &= &\mathbf{x_i}(t)+f(Q,{x}_i,a) \\
 f(Q,{x}_i,a) & = &\scalebox{0.90}[1]{$\{\hat{x}_i|\max [Q(x_i,a)]~~~for~~ \forall a\in A \}$}
\label{final_update_2}
\end{eqnarray}
where $ \mathbf{\hat{x}_i}$ is a particular position vector obtained from $f(Q,{x}_i,a)$ in which action $a$ is taken since it has the highest utility value, $Q$, returned. The main aim of each individual/particle is to learn from the experiences gained once each receives the reward produced by reinforcement mechanism with crediting the action rightly taken and punishing the wrongly taken ones. This learning property to be incrementally developed by each particle will succeed to a well-designed collective behaviour.
\subsection{Reinforcement Mechanism}\label{reinf_mech}
As clearly indicated before, reinforcement mechanism plays the crucial role in furnishing particles with learning capabilities. It remains as an independent monitoring mechanism to assess the actions taken by the particles and supply them with reinforcing payoff grades. It is usually implemented in a Reward Function, which is defined as follows:
\begin{equation}
\mathbf{R}:S\times A\;\longrightarrow \mathbb{R}
\end{equation}
The reward function is implemented to consider the situation with a particular state, $x$, applied with action $a$, whether it is or not the correct action taken. A reward, $r$, will be produced as the assessment level for the situation. Thus, an efficient reward function will be developed based on the problem domain.

\section{Scanning disastrous area with swarm of learning agents}\label{implementation}

\begin{figure}[tb]
    \begin{center}
     \begin{tabular}{c}
		\includegraphics[width = 8cm]{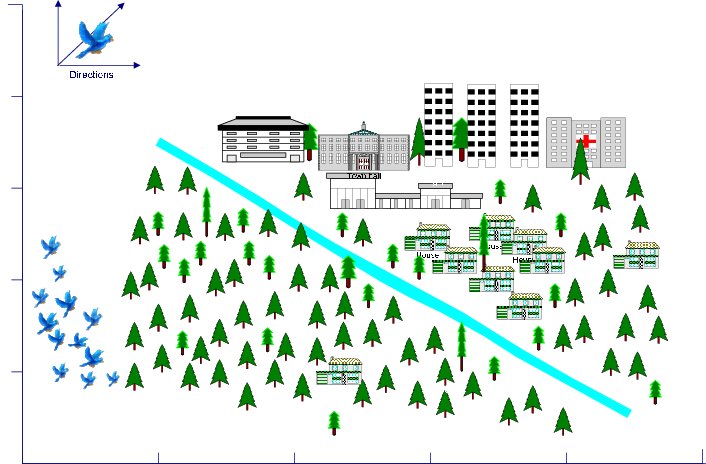}
     \end{tabular}
  \end{center}
\caption{\small A typical urban landscape}
\label{fig:landscape}
\end{figure}

\begin{figure}[b]
    \begin{center}
     \begin{tabular}{c}
		\includegraphics[width = 6 cm]{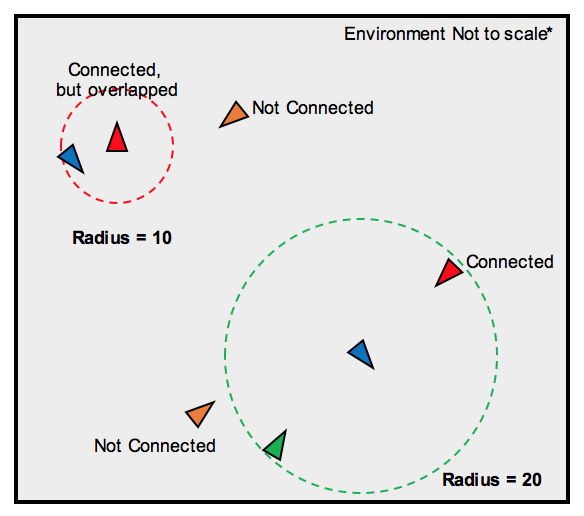}
     \end{tabular}
  \end{center}
\caption{\small Connecting individuals via distance }
\label{fig:connect}
\end{figure}

This problem case is adopted to illustrate the implementation of collective intelligence achieved using the multi agent learning algorithm proposed in this study, which is built up through embedding Q learning within particle swarm optimisation algorithm. Fig~\ref{fig:landscape} illustrates a simple scenario in which a typical piece of land combining rural and urban areas to be scanned by a swarm of learning agents. Suppose that such an area subjected to some disasters is required to be scanned for information collection purposes. A flock of artificial birds (swarm of UAVs); each is identified as a particle and furnished with a list of actions to take while moving around the area in collaboration with other peer particles. Each particle is enabled to learn via the Q learning implemented for this purpose and being trained how to remain connected with the rest of the swarm. The logic is implemented to identify if a particle is collaborating or not as demonstrated in Fig.~\ref{fig:connect}. 

Two possible cases are illustrated in Fig.~\ref{fig:connect}. As indicated, teams of particles (flocks of birds) can remain interconnected for collaboration if each is sufficiently close to another peer particle, which is measured with Euclidean distance that is particularly calculated in a circle-centric way. A particle is considered connected if remains within a circle with particular radius, but will be out of connection if remains out of the circle of that radius. In addition, the particles are expected not to approach to each other beyond a certain distance, then they will also be counted not well-collaborating since they overlap and cause wasting resources. The main idea behind this algorithm is to train the particles not to fall apart and not to overlap, either. That is the main objective to achieve.  

\subsection{Embedding Q learning within each particle}\label{particle}
Since the swarm intelligence framework preferred in this study is PSO, each individual to form up the swarm will be identified as a particle as is in particle swarm optimisation. Let $M$ be the size of the swarm, where $M$ particles are created to form up the swarm; each has a 2-dimensional position vector, $\mathbf{x_i}=\{x_{1,i}, x_{2,i}|~i=1,...,M\}$, because the defined area is 2-dimensional and each particle will simply move forward and/or backward, vertically and/or horizontally. For simplification purposes, each particle is allowed to move with selecting one of predefined actions, where each action is defined as a step in which the particle can chose the size of the step only. Using the same notation as Q learning, the size of set of actions is $A$, which includes forward and backward short, middle and long size steps. Hence, a particle can move forward and backward with selecting one of these six actions. Let $\mathbf{\Delta}=\{\delta_{j}|j=1,...,A\}$ be the set steps including both forward and backward ones, which a particle is able to take as part of the action it wants to do. Once an action is decided and taken, the position of the particle will change as much as:-
\begin{eqnarray}
 f(Q,{x}_i,a) & = &\pi\delta_{j}
 \label{step}
\end{eqnarray}
where $\pi$ is a probability calculated based on position and possible move of neighbouring particles. Substituting equation (\ref{step}) within equation (\ref{final_update_2}), the new position of the particle under consideration is determined.  Here, the neighbourhood is considered as the other peer particles that has connectivity with the one under consideration, which is determined based on the distance in between. Let $N_i\in M$ be the set of neighbouring peer particles (agents) of $ith$ particle, which is defined as:-
 \begin{eqnarray}
N_i & = &\{\mathbf{x_k}|~\epsilon > d(\mathbf{x_i},\mathbf{x_k})\}~~~\forall k \in M
\end{eqnarray}
where $d(\mathbf{x_i},\mathbf{x_k})$ is calculated as a Euclidean distance and $\epsilon$ is the maximum distance, (the threshold), between two peer particles set up to remain connected. Once a particle moved as a result of the action taken, the reinforcement mechanism, the reward function in another name, assesses the decision made for this action considering the previous state of the particle before transition and the resulted position of neighbouring peer particles.

\begin{eqnarray}
 r & = &\left\{
 \begin{array}{lr}
 100,&\scalebox{0.75}[1]{$\mbox{if}~~\sum_{k=1}^{N_i} d(\mathbf{x_i}, \mathbf{x_k}) =  N_i\epsilon$}\\\\
 \scalebox{0.75}[1]{$N_i\epsilon-\sum_{k=1}^{N_i} d(\mathbf{x_i},\mathbf{x_k}),$}&\scalebox{0.75}[1]{$\mbox{if}~~ N_i\epsilon > \sum_{k=1}^{N_i} d(\mathbf{x_i}, \mathbf{x_k})$} \\\\
 -100,&\scalebox{0.75}[1]{$\mbox{if}~~\sum_{k=1}^{N_i}d(\mathbf{x_i},\mathbf{x_k})\leq 0$}\\
 \end{array}\right.
 \label{r_function}
\end{eqnarray}

$\epsilon$ is also the maximum sensing distance of each particle in which the particles allowed to be apart and connected. The reward is mainly calculated based on the total distance from the particle to its neighbouring particles. If there is no neighbouring particle determined, which means the particle has lost connection, then it will be punished with $-100$ negative reward. If there is still connection but is less then  $N_i\epsilon $, then the negative reward will be as much as calculated in the second option of equation (\ref{r_function}). If the total distance from its position to all other neighbouring particles equals to $N_i\epsilon $, then that deserves the whole reward, which is $100$.

\section{Experimental Results}\label{results}
%\subsection{Introduction} 
This section presents experimental results to demonstrate a proof-of-concept Q learning algorithm works to help particles (agents) self-train towards building a collaboration and behave as an swarm member.  The aim is also to revise and analyse how the whole study turned out, judging whether the final implementation adhered to the expectations pre-set up. The algorithm has been implemented for a number of swarm sizes using an agent-based simulation tool called NetLogo~\cite{Wilensky_15}. 

For a successful evaluation, an agile approach has been adopted to run the study through iterations in which the study has been incremented bit-by-bit. As per the approach various elements were considered ranging from the algorithm itself to the methods, techniques and tools used, analysing what each component did well and what could be done better. One way to get an insight into the level of success of each aspect of the study project is to imagine starting the same project fresh whilst retaining all current knowledge and contemplating what elements would be kept and what would be changed, and whether these changes could lead to an improved implementation. 
\subsection{Approximation and Evaluation}
Throughout the project, the initial iteration was to start the study to find out a way to embed Q learning into PSO, which has been achieved in the previous sections as explained.  The ultimate aim is to show that both algorithm work hand-by-hand to achieve a a swarm of learning agents which collaborate for collective behavior/intelligence. This aim is not so black and white, but has many grey areas involved. This is because the algorithm is not just looking at the speed of convergence for example which could very easily be answered as to whether improvement has been made. Rather, the algorithm is subject to in depth observation as to whether the particles are behaving correctly, which in itself has intricacies that require close inspection.

In increment two, the goal was to get each particle to essentially be "reactive" each other in a real world environment. So if one particle moved, the others which are also moving simultaneously would need to take their fellow particles movement as well as their own into consideration and react accordingly so that they are always within proximity of their neighbours. This proximity prevents a particle from invading its neighbours space whilst also not allowing it to drift too far out of the radius, if it does either of these it will get punished whereas if it stays the "perfect" distance away, it will get the maximum reward. 

This approach in theory would allow collaborative learning to occur as particles gain the knowledge of the correct expected behaviour. This brings the question as to whether this was successful or not to which the results suggest it very much was. Each participating particle was actively reacting to the movement of its peers, and with the reinforcement mechanic, they were learning which actions would be best to take with each iteration. In this iteration, two swarms are created for which one was working with embedded Q , learning (will be presented with the acronym of M-QL here-forth) and the other was run with a standard PSO.

\begin{figure}[!tbp]
 % \centering
  \begin{minipage}[b]{0.45\linewidth}
    \includegraphics[width=\linewidth]{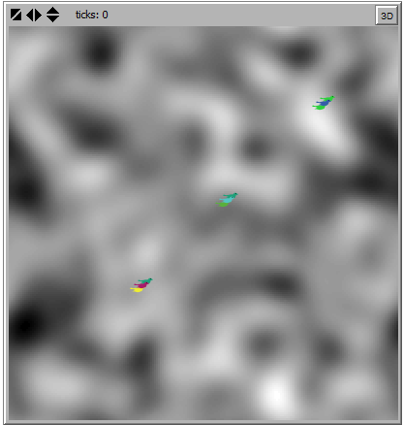}
    \subcaption{\small Initial positions for M-QL}
    \label{fig:init_Q}
  \end{minipage}
  \hfill
  \begin{minipage}[b]{0.45\linewidth}
    \includegraphics[width=\linewidth]{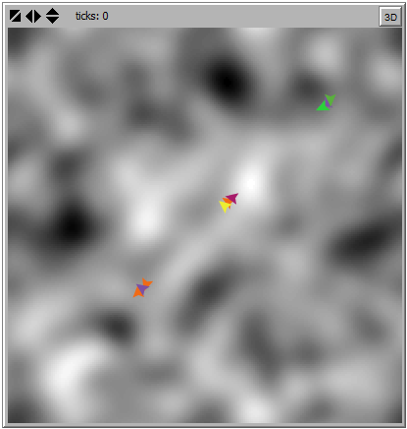}
    \subcaption{\small Initial positions for PSO}
    \label{fig:init_pso}
  \end{minipage}
  \caption{\small Initialised particles' positions for both algorithms; multi-agent Q learning and particle swarm optimisation}
  \label{fig:init}
\end{figure}

\begin{figure}[!tbp]
  \centering
  \begin{minipage}[b]{0.45\linewidth}
    \includegraphics[width=\linewidth]{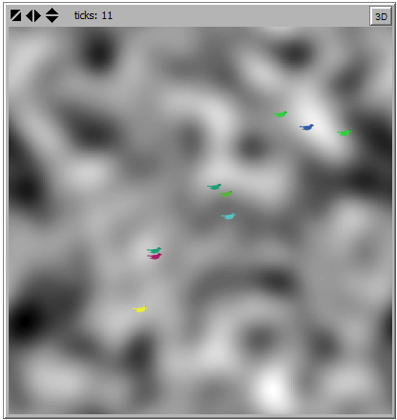}
    \subcaption{\small  After 10 iterations with m-QL}
    \label{fig:ticks_10_Q}
  \end{minipage}
  \hfill
  \begin{minipage}[b]{0.45\linewidth}
    \includegraphics[width=\linewidth]{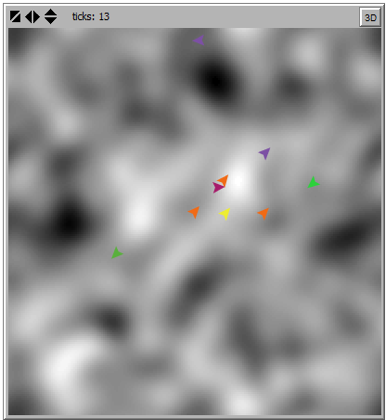}
    \subcaption{\small After 10 iterations with PSO}
    \label{fig:ticks_10_pso}
  \end{minipage}
  \begin{minipage}[b]{0.45\linewidth}
    \includegraphics[width=\linewidth]{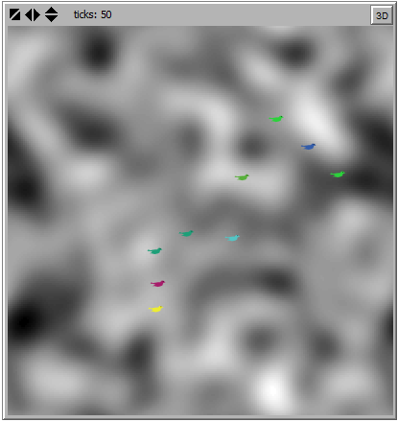}
    \subcaption{\small After 50 iterations with m-QL}
    \label{fig:ticks_50_Q}
  \end{minipage}
  \hfill
  \begin{minipage}[b]{0.45\linewidth}
    \includegraphics[width=\linewidth]{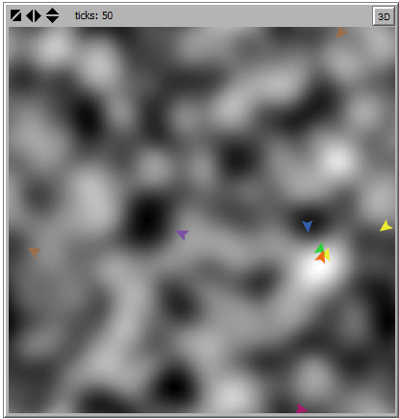}
    \subcaption{\small After 50 iterations with PSO}
    \label{fig:ticks_50_pso}
  \end{minipage}
  
  \begin{minipage}[b]{0.45\linewidth}
    \includegraphics[width=\linewidth]{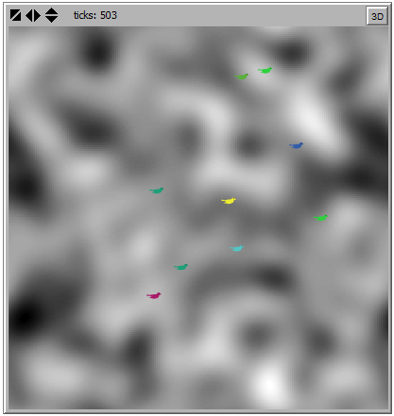}
    \subcaption{\small After 500 iterations with m-QL}
    \label{fig:ticks_500_Q}
  \end{minipage}
  \hfill
  \begin{minipage}[b]{0.45\linewidth}
    \includegraphics[width=\linewidth]{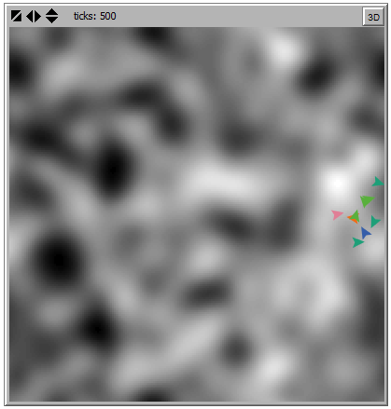}
    \subcaption{\small After 500 iterations with PSO}
    \label{fig:ticks_500_pso}
  \end{minipage}
  \caption{A set of comparative results to demonstrate the behaviours of the learning algorithm versus PSO }
  \label{fig:ticks}
\end{figure}

The experimentation is organised to start with the initial swarms as seen in Fig.~\ref{fig:init} and then the swarms are incremented through iteration as presented in Fig.~\ref{fig:ticks}, where the behaviors of both swarms, learning swarm and PSO swarm, after 10, 50 and 500 iterations, respectively.  
%\begin{strip}

Fig.~\ref{fig:init} and Fig.~\ref{fig:ticks} illustrate the stark difference in which the particles move with a reinforced incentive. Whereas PSO is essentially solely designed to iteratively move particles towards their best value, the addition of the Q-learning proximity measure prevents such erratic movement. Of course over 500 iterations, some movement is going to occur as particles will rarely be hitting their "perfect" +100 reward movements, but as is visible in Fig.~\ref{fig:ticks_500_Q}, each particle is connected to at least one other in a feasible proximity. In this instance, which does not always happen, the clusters have merged together indirectly causing one large network. This is fine and can be expected to happen on occasion as through individual incremental movement through the environment, particles are going to move into the consideration radius of other particles subsequently inheriting them into their peer particle Q-mechanic. This is something that in a real world situation might need to be prevented if clusters are required to remain in that native cluster, however in this simple environment, with no mechanic to prevent it, it is acceptable.

%\onecolumn
%\end{strip}
%\twocolumn
As can be observed from Fig.~\ref{fig:ticks}, the particles of the swarm, learning with M-QL, can demonstrate connectivity among themselves via having a connecting distance from one another while the swarm running PSO approximates to a particular value, where all particles nearly come to overlapping positions. In fact, the behaviours of the particles in Fig.~\ref{fig:ticks_10_Q}, ~\ref{fig:ticks_50_Q}, ~\ref{fig:ticks_500_Q} clearly indicates that the individual particles keeping distance neither much falling apart nor remaining too close to one another, while the number of iterations increases the distances become more fitting as Fig.~\ref{fig:ticks_10_Q} shows some particles are still too close to each other, but, Fig.~\ref{fig:ticks_500_Q} indicates a better positioning. On the other hand, Fig.~\ref{fig:ticks_10_pso}, ~\ref{fig:ticks_50_pso}, ~\ref{fig:ticks_500_pso} demonstrate how particles approximate to a targeted value without considering any having any distance among one another. More iterations help individual particles getting closer and taking overlapping positions more and more.    

The results of increment two confirm that particles are at least capable of learning both in the individual sense and the subsequent group sense. Although this is a fundamentally basic example of learning, it acts as a basis which can be built on in various ways. From running various parameter configurations in earlier experimentations, it was observed that the particles choose the correct action to take in relation to the proximity as this showed they had learned which action would benefit them the most, which also showed the components of state and action were working correctly.
However as estimated, the workings of the increment are not perfect as whilst through the first phase of iterations the clusters seem to keep a good proximity with particles clearly getting negative rewards for interfering with their counterparts.

It is observed that as the episode gets near the last 40\% of iterations on average, particles visibly begin to drift out of the neighbourhood, and once a particle loses connection, there is no mechanic to get the particle back into a neighbourhood radius, only luck can allow this to happen. Of course if this incident happened in a real world environment, the results could be extremely costly. Therefore this issue is left to further research in the future through potential reinforced path-finding or efficient search algorithms. If a particle could find its way back into a neighbourhood, the algorithm could be much more efficient and realistically deployable in a real world domain.

\subsection{Individuals' learning behaviour}
The performance of learning particles was another aim of this research. For observing individual learning performance, three particles are taken under observation over 100 iterations.  Due to the limitations of NetLogo, each simulation in this regard is physically observed from start to end, for each iteration, particles are individually judged whether each has made a good decision or a bad decision, good decision means taking the correct action and getting positive reward while bad decision indicates taking wrong actions and receiving negative rewards (penalty). 

For quantification, a good decision will be dictated by a particle moving in such a way it does not get too close to a fellow particle and does not drift outside of the radius either. This brings into question the issue of synchronisation. The synchronisation problem occurs when two particles move at the same time which can cause two particles to "choose" to move closer to each other at the same time or move further away, thus causing a bad decision.

% \begin{figure}%[!tbp]
%   \centering
%   \begin{minipage}[b]{0.15\textwidth}
%     \includegraphics[width=\textwidth]{picts/ind-1.png}
%     \subcaption{\small Particle 1}
%     \label{ind_1}
%   \end{minipage}
%   \hfill
%   \begin{minipage}[b]{0.15\textwidth}
%     \includegraphics[width=\textwidth]{picts/ind-2.png}
%     \subcaption{\small Particle 2}
%     \label{ind_2}
%   \end{minipage}
%   \begin{minipage}[b]{0.15\textwidth}
%     \includegraphics[width=\textwidth]{picts/ind-3.png}
%     \subcaption{\small Particle 3}
%     \label{ind_3}
%   \end{minipage}
%   \caption{A flock of three birds considered for individual learning to collaborate}
%   \label{fig:ind}
% \end{figure}

It also must be noted that when a particle moves out of a radius, it has no real method of finding its neighbourhood again and therefore it becomes a flat-line of bad decisions on the graph.

\begin{figure}[!tbp]
  \centering
  \begin{minipage}[b]{0.35\textwidth}
    \includegraphics[width=\textwidth]{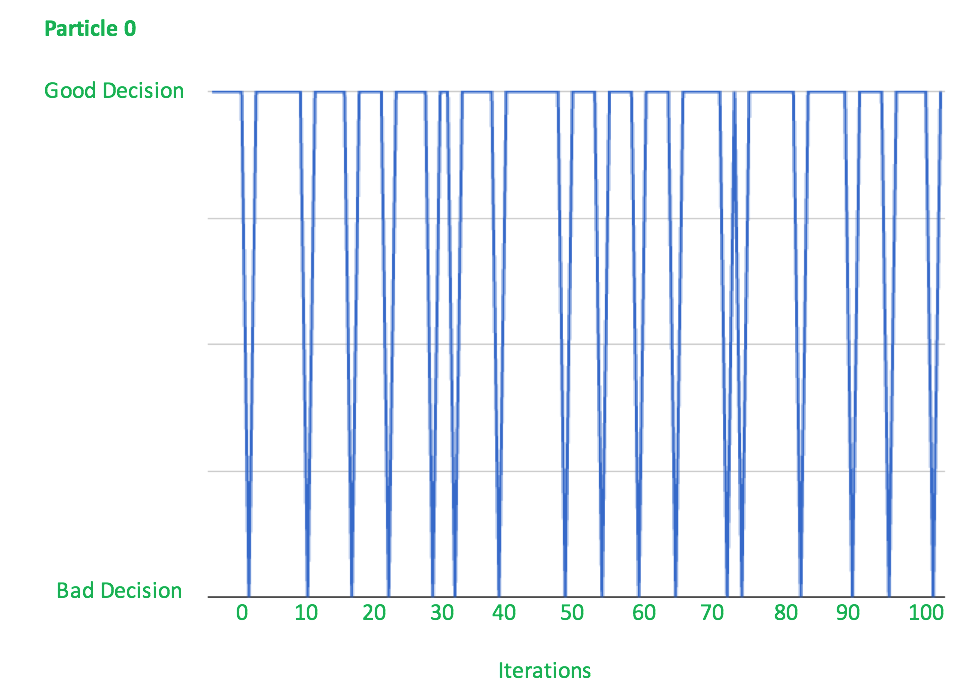}
    \subcaption{\small Particle 0}
    \label{fig:indv_1}
  \end{minipage}
  \hfill
  \begin{minipage}[b]{0.35\textwidth}
    \includegraphics[width=\textwidth]{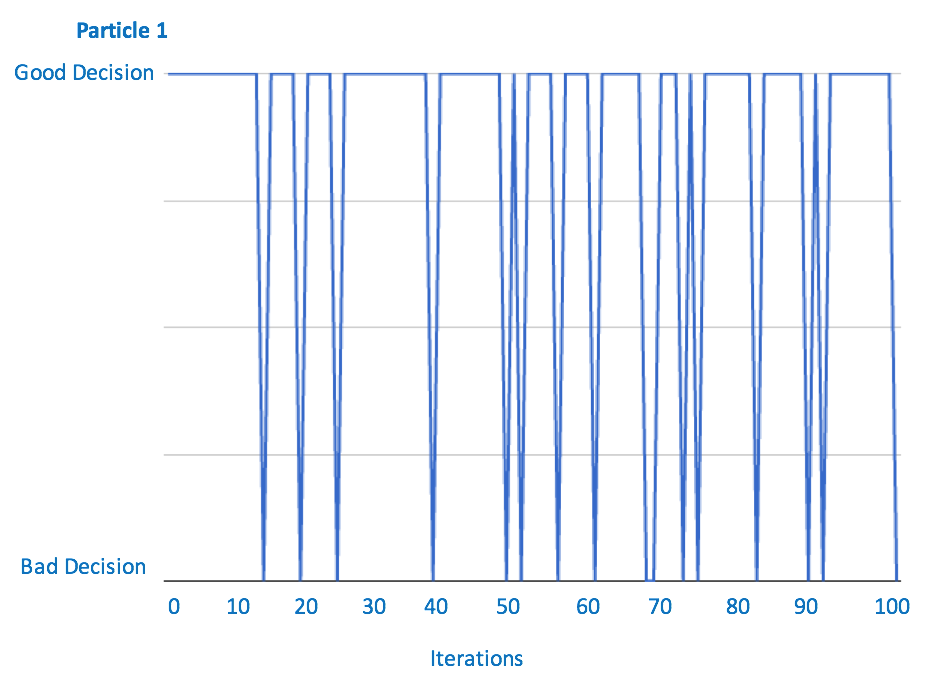}
    \subcaption{\small Particle 1}
    \label{fig:indv_2}
  \end{minipage}
  \begin{minipage}[b]{0.35\textwidth}
    \includegraphics[width=\textwidth]{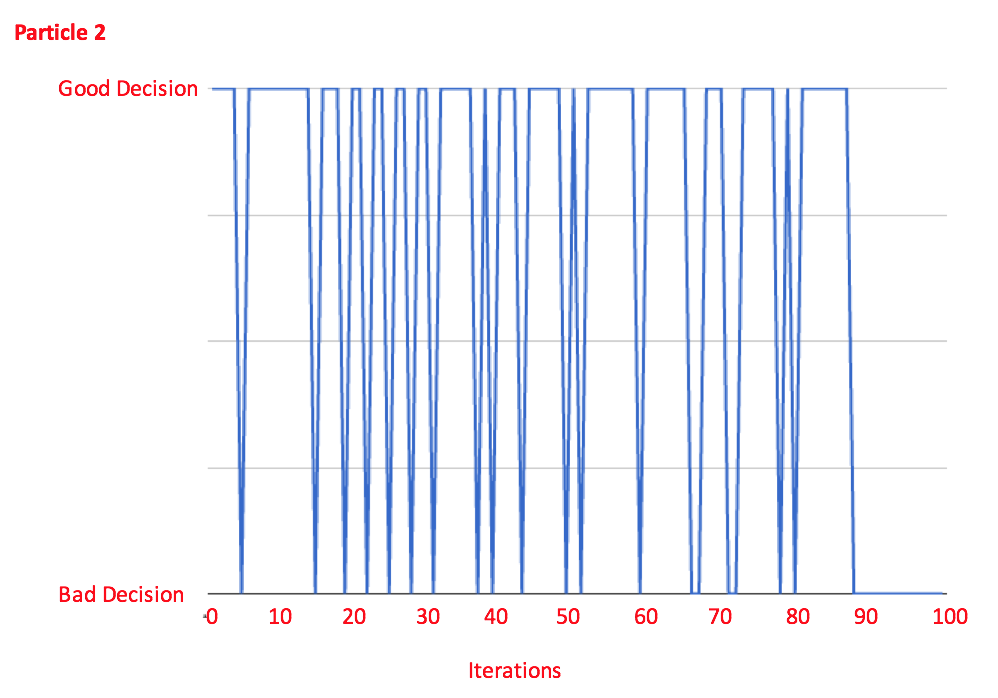}
    \subcaption{\small Particle 2}
    \label{fig:indv_3}
  \end{minipage}
  \caption{Learning behaviours of the three particles. }
  \label{fig:indv}
\end{figure}

The results of these graphs show that for the most part, the correct decision is usually made, which shows the hybridised algorithm does work. As mentioned before, an issue occurs when two particles move at the same time because they simply cannot predict what their fellow particles will do which causes proximity problems. This could simply be rectified by having certain particles in the topology moving in "turns". If one particle is not moving on a turn, this would allow the other particles to successfully move closer or further away from it without any conflict.

As can be seen in the graph for particle 2~(Fig.~\ref{fig:indv_3}), it started to make bad decisions for around 10 iterations which also continued after. This was because it drifted out of the radius of its topology and has no method of getting back into it. This is another problem that I believe could be simply rectified, if a technique was implemented to allow this lost particle to re-find or find another topology. Even a random search method would give decent results if the landscape was well populated.
I also changed the reward value from 5 to 0.5 but this had little to no difference shown in the graphs, as the subsequent discount factor and learning rate don’t get enough of a chance to have a real impetus on the results. The changing of these parameters would be much more effective over longer iterations such as 500-1000, however the output configuration of NetLogo makes any further analysis other than observation hard.

\section{Conclusions}\label{conclusions}
This paper presents a proof-of concept study for demonstrating the viability of building collaboration among multiple agents through standard Q learning algorithm embedded in particle swarm optimisation. A number of particles furnished with Q learning has been subjected to self training to act as members of a swarm and produce collaborative/collective behaviours. Following introducing the algorithmic foundation and structures, an experimental study is conducted to demonstrate that the formulated algorithm produces results supporting the aimed behaviours of the algorithm. The results are produced with very simplistic assumptions, where further enhancements require further extensive theoretical and experimental studies.

 %\bibliographystyle{ieeeconf}
 %\bibliography{emn.bib, MyBib.bib, ant.bib}

\begin{thebibliography}{99}
\bibitem{Alechina_Logan_10} N. Alechina, and B. Logan: Computationally grounded account of belief and awareness for AI agents, In Proceedings of The Multi-Agent Logics, Languages, and Organisations Federated Workshops (MALLOW 2010), Lyon, France, August 30 - September 2, 2010, volume CEUR-WS 627.
\bibitem{Ayhan_13} M.B. Ayhan, M.E. Aydin, and E. Oztemel: A multi-agent based approach for change management in manufacturing enterprises. Journal of Intelligent Manufacturing, 26 (5), 2015, pp. 975-988.
\bibitem{Aydin_et_al_11} M.E. Aydin,  N. Bessis, E. Asimakopoulou, F. Xhafa, and  J. Wu: Scanning Environments with Swarms of Learning Birds: A Computational Intelligence Approach for Managing Disasters. In IEEE International Conference on Advanced Information Networking and Applications (AINA), 2011, pp. 332-339. 
\bibitem{Aydin_10} M. E. Aydin: Coordinating metaheuristic agents with swarm intelligence, Journal of Intelligent Manufacturing, 23, 4 (August 2012), 991-999.
\bibitem{Aydin_et_al_09}M. E. Aydin, R. Kwan, C. Leung, and J. Zhang: Multiuser scheduling in hsdpa with particle swarm optimization, in Applications Of Evolutionary Computing, Proceedings, Giacobini, ed., vol. 5484 of Lecture Notes In Computer Science, 2009, pp. 71-80.
\bibitem{Aydin_07} M. Aydin: Metaheuristic agent teams for job shop scheduling problems. In Holonic and Multi-Agent Systems for Manufacturing, 2007, pp. 185-194.
\bibitem{Bradtke_96}J. Bradtke, A. G. Barto, and P. Kaelbling: Linear least-squares algorithms for temporal difference learning, in Machine Learning, 1996, pp. 22-33. 
\bibitem{Bull_05}L. Bull: Two Simple Learning Classier Systems, in Foundations of Learning Classier Systems, L. Bull and T. Kovacs, eds., no. 183 in Studies in Fuzziness and Soft Computing, Springer-Verlag, 2005, pp. 63-90. 
\bibitem{Bull_et_al_05}L. Bull and T. Kovacs, eds., Foundations of Learning Classier Systems, vol. 183 of Studies in Fuzziness and Soft Computing, Springer, 2005.
\bibitem{Claus_98}C. Claus and C. Boutilier: The dynamics of reinforcement learning in cooperative multiagent systems, in In Proceedings of National Conference on Artificial Intelligence (AAAI-98, 1998, pp. 746-752.
\bibitem{Colorni_et_al_94}A. Colorni, M. Dorigo, V. Maniezzo, and M. Trubian: Ant system for job-shop scheduling. Belgian Journal of Operations Research, Statistics and Computer Science (JORBEL), 34(1) 1994, pp. 39-53.
\bibitem{Dong_16}X. Dong: Consensus Control of Swarm Systems, In Formation and Containment Control for High-order Linear Swarm Systems, 2016, pp. 33-51. Springer Berlin Heidelberg.
\bibitem{Eberhart_95}R. Eberhart and J. Kennedy: A new optimizer using particle swarm theory, in Proc. of the 6th Int. Symposium on Micro-Machine and Human Science, 1995, pp. 39 - 43. 
\bibitem{Foerster_16} J. Foerster, Y.M.Assael, N. de Freitas, and S. Whiteson: Learning to communicate with deep multi-agent reinforcement learning, In Advances in Neural Information Processing Systems 2016, pp. 2137-2145.
\bibitem{Gath_15}M. Gath:Optimizing Transport Logistics Processes with Multiagent Planning and Control. PhD Thesis, 2015, published by Springer Fachmedien Wiesbaden; 2016 Jul 11.
\bibitem{Hammami_05}. M. Hammami, and K. Ghediera: COSATS, X-COSATS: Two multi-agent systems cooperating simulated annealing, tabu search and X-over operator for the K-Graph Partitioning problem, Lecture Notes in Computer Science 3684, 2005, p. 647-653.
\bibitem{Hercog_13} L. M. Hercog:Better manufacturing process organization using multi-agent self-organization and co-evolutionary classifier systems: The multibar problem. Appl. Soft Comput. 13(3),2013, pp. 1407-1418. 
\bibitem{Iima_09} H. Iima and Y. Kuroe: Swarm reinforcement learning algorithm based on particle swarm optimization whose personal bests have lifespans, in Neural Information Processing, C. Leung, M. Lee, and J. Chan, eds., vol. 5864 of Lecture Notes in Computer Science, Springer Berlin / Heidelberg, 2009, pp. 169-178.
\bibitem{Kazemi_09} A. Kazemi, M. F. Zarandi, and  S. M. Husseini:  A multi-agent system to solve the production–distribution planning problem for a supply chain: a genetic algorithm approach. The International Journal of Advanced Manufacturing Technology, 44(1-2), 2009, pp.180-193.
\bibitem{Kennedy_97}J. Kennedy and R. C. Eberhart: A discrete binary version of the particle swarm algorithm," 1997 IEEE International Conference on Systems, Man, and Cybernetics. Computational Cybernetics and Simulation, Orlando, FL, 1997, pp. 4104-4108. 
\bibitem{Kennedy_01}J. Kennedy, R. Eberhart, and Y. Shi.: Swarm Intelligence, Morgan Kaufmann, San Mateo, CA, USA, 2001.
\bibitem{Kok_04}J. R. Kok and N. Vlassis: Sparse cooperative q-learning, in Proceedings of the International Conference on Machine Learning, ACM, 2004, pp. 481-488. 
\bibitem{Kolp_06} M. Kolp, P. Giorgini, and J. Mylopoulos: Multi-agent architectures as organizational structures, Autonomous Agents and Multi-Agent Systems, 13, 2006, pp. 3-25.
\bibitem{Kouider_12} A. Kouider,  and  B. Bouzouia: Multi-agent job shop scheduling system based on co-operative approach of idle time minimisation. International Journal of Production Research, 50(2), 2012, pp.409-424.
\bibitem{Kouvaros_16} P. Kouvaros and A. Lomuscio: Parameterised verification for multi-agent systems, Artificial  Intelligence, 234, C (May 2016), pp. 152-189.
\bibitem{Meng_08}Y. Meng: Recent Advances in Multi-Robot Systems, I-Tech Education and Publishing, 2008, ch. Q-Learning Adjusted Bio-Inspired Multi-Robot Coordination, pp. 139-152. 
\bibitem{Mohebbi_12} S. Mohebbi, and R. Shafaei: E-Supply network coordination: The design of intelligent agents for buyer-supplier dynamic negotiations. Journal of Intelligent Manufacturing 23, (2012), pp.375-391.
\bibitem{Panait_05}L. Panait and S. Luke: Cooperative multi-agent learning: The state of the art. Autonomous agents and multi-agent systems, 11(3),2005, pp.387-434.
\bibitem{Poli_07}R. Poli, J. Kennedy, and T. Blackwell: Particle swarm optimization. Swarm Intelligence, 1 2007, pp. 33-57. 
\bibitem{Sutton_98}R. S. Sutton and A. G. Barto: Reinforcement Learning: An Introduction, MIT Press, Cambridge, MA, USA, 1998. 
\bibitem{Tasgetiren_07}M. Tasgetiren, Y. Liang, M. Sevkli, and G. Gencyilmaz: Particle swarm optimization algorithm for makespan and total flow-time minimization in permutation flow-shop sequencing problem. European Journal of Operational Research, 177(3) 2007, pp. 1930-1947.
\bibitem{Tesauro_92}G. Tesauro: Practical issues in temporal difference learning, in Machine Learning, 1992, pp. 257-277. 
\bibitem{Tsitsiklis_94}J. N. Tsitsiklis and R. Sutton: Asynchronous stochastic approximation and Q-learning, in Machine Learning, 1994, pp. 185-202. 
\bibitem{Vazquez_05} J. Vazquez-Salceda, V. Dignum, and F. Dignum: Organizing multi-agent systems, Autonomous Agents and Multi-Agent Systems, 11, 2005, pp. 307-360.
\bibitem{Watkins_89}C. Watkins: Learning from delayed rewards, PhD thesis, Cambridge University, 1989. 
\bibitem{Watkins_92}C. Watkins and P. Dayan: Technical note: Q-learning. Machine Learning, 8, 1992, pp. 279- 292. 
\bibitem{Wilensky_15} U. Wilensky and W. Rand: An introduction to agent-based modeling: Modeling natural, social and engineered complex systems with NetLogo, MIT Press, Cambridge, 2015.

\end{thebibliography}
% \printbibliography
\end{document}